\documentclass{article}
\usepackage{ijcai21}

\usepackage{times}

\usepackage{soul}
\usepackage{url}
\usepackage[hidelinks]{hyperref}
\usepackage[utf8]{inputenc}
\usepackage[small]{caption}
\usepackage{graphicx}
\usepackage{amsmath}
\usepackage{booktabs}
\usepackage{subfigure}
\usepackage{amsmath}
\usepackage{amsthm}
\usepackage{multirow}
\usepackage{amsfonts}

\title{IC Networks: Remodeling the Basic Unit for Convolutional Neural Networks}

\author{
\author{
	Junyi An$^1$
	Fengshan Liu$^1$
	Jian zhao$^2$
	Furao shen$^1$\footnote{Contact Author}\\
	$^1$Department of Computer Science and Technology, Nanjing University, Nanjing, China\\
	$^2$School of Electronic Science and Engineering, Nanjing University, Nanjing, China\\
	\texttt{ \{junyian, liufengshan\}@smail.nju.edu.cn }\\
	\texttt{ \{frshen, jianzhao\}@nju.edu.cn }
}
}

\begin{document}

\maketitle

\begin{abstract}

Convolutional neural network (CNN) is a class of artificial neural networks widely used in computer vision tasks. Most CNNs achieve excellent performance by stacking certain types of basic units. In addition to increasing the depth and width of the network, designing more effective basic units has become an important research topic. Inspired by the elastic collision model in physics, we present a general structure which can be integrated into the existing CNNs to improve their performance. We term it the “Inter-layer Collision” (IC) structure. Compared to the traditional convolution structure, the IC structure introduces nonlinearity and feature recalibration in the linear convolution operation, which can capture more fine-grained features. In addition, a new training method, namely weak logit distillation (WLD), is proposed to speed up the training of IC networks by extracting knowledge from pre-trained basic models. In the ImageNet experiment, we integrate the IC structure into ResNet-50 and reduce the top-1 error from $22.38\%$ to $21.75\%$, which also catches up the top-1 error of ResNet-100 ($21.75\%$) with nearly half of FLOPs. 
\end{abstract}

\section{Introduction}
Convolutional neural networks (CNNs) have made great achievements in the field of computer vision. The success of AlexNet \cite{krizhevsky2012imagenet} and VGGNet \cite{simonyan2014very} shows the superiority of deep networks, leading to a trend of building larger and deeper networks. However, this method is inefficient in improving network performance. On the one hand, increasing the depth and width brings a huge computational burden and causes a series of problems, such as model degradation. On the other hand, because the relationship between different hyper-parameters is complicated, the increased number of hyper-parameters makes it more difficult to design deep networks. Therefore, the research focus in recent years has gradually shifted to improving the representation ability of basic network units in order to design more efficient CNN architectures.

The convolutional layer is a basic unit of CNN. By stacking a series of convolutional layers together with non-linear activation layers, CNNs are able to produce image representations that capture abstract features and cover global theoretical receptive fields. For each convolutional layer, the filter can capture a certain type of feature in the input. However, not all features contribute to a given task. Recent studies have shown that the network can obtain more powerful representation ability by feature recalibration, which emphasizes informative features and suppresses less useful ones~\cite{bell2016inside,hu2018squeeze}. Besides, there is evidence that introducing the non-linear kernel method in the convolutional layer can improve the generalization ability of the network \cite{wang2019kervolutional,zoumpourlis2017non}. However, kernel methods may cause overfitting by complicating the networks and introducing a high amount of calculation. 

We argue that the convolutional layer can also benefit from a simple non-linear representation, and propose a structure of introducing a non-linear operation in the convolutional layer, which also performs feature recalibration to enhance the representation ability of convolutional layers. Starting from the most basic neural network structure where a linear transformation and a non-linear activation function are applied to the input successively, we use a non-linear operation to optimize the representation of the linear part. The proposed structure divides the input space into multiple subspaces which can represent different linear transformations, providing more patterns for succedent activation function. We call this structure the \emph{inter-layer collision} (IC) neuron, since it is inspired by the elastic collision model that we use to mimic the way information is transmitted between adjacent layers.
% Through it, the input space can be divided into multiple subspaces which can represent different linear transformations, !providing more patterns for succedent activation function. Therefore, the proposed structure can learn the abstract non-linear representation. We called this structure the Inter-layer Collision (IC) neuron, since it is inspired by an elastic collision model in physics which make the feedforward propagation like the collision between neurons in adjacent layers.

\begin{figure}
    \centering
    \includegraphics[scale=0.2]{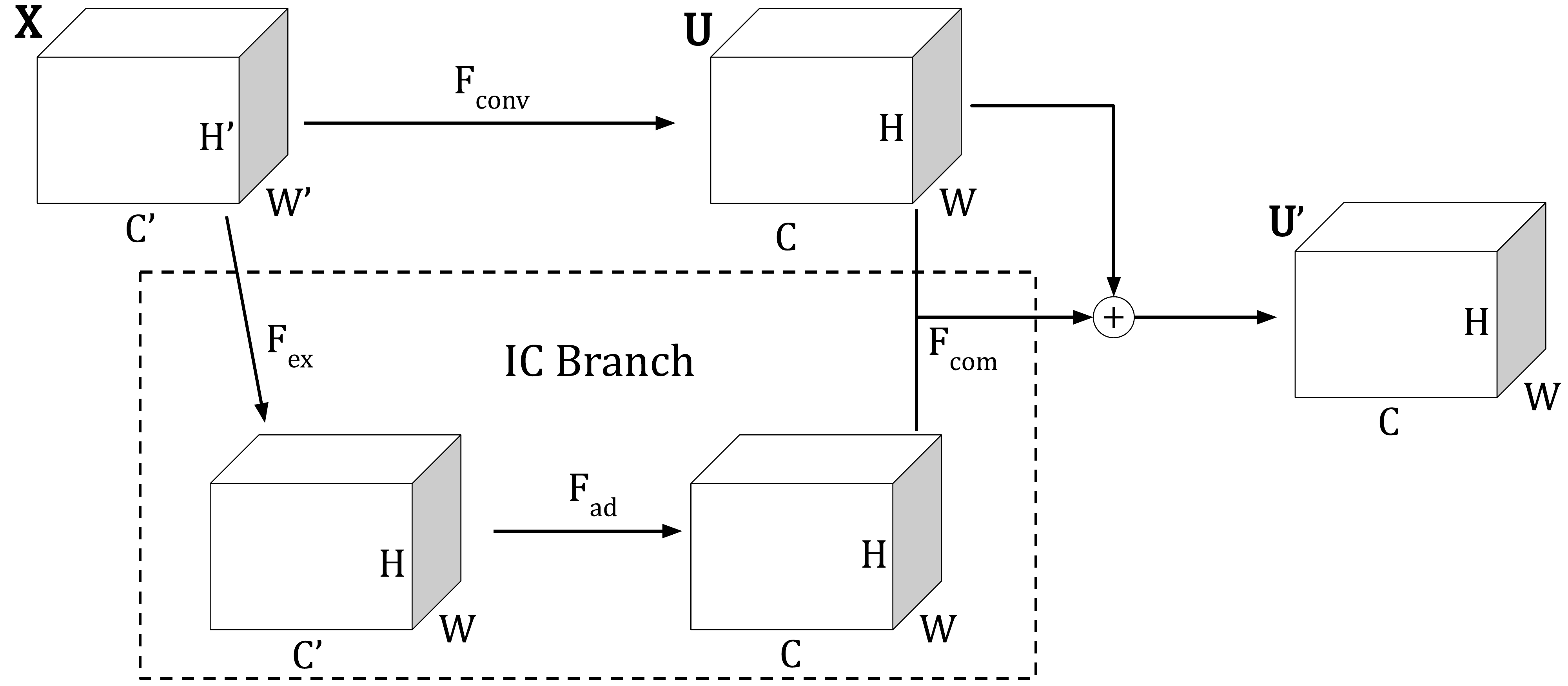}
    \caption{The structure of a complete IC layer.}
    \label{fig1}
\end{figure}

We build the IC convolutional layer by combining the IC neuron with the convolution operation. The structure of the IC layer is depicted in Figure~\ref{fig1}, where the term $F_{conv}$ denotes the convolution operation mapping the input $\mathbf{X}$ to the feature maps $\mathbf{U}$. An IC branch used to divide the input space can be easily combined with the $F_{conv}$ branch. When the input $\mathbf{X}$ passes though the IC branch, the $F_{ex}$ operation first extract local features by aggregating input features in local regions, which represents the local distribution of channel-wise input features. Then, local features are recalibrated by the adjustment operation $F_{ad}$ to increase flexibility of representation. Finally, the local features pass though a $F_{com}$ operation which combines two branches to generate extra linear transformation. The final output $\mathbf{U}'$ of the IC layer with the same spatial dimensions ($H \times W$) as $\mathbf{U}$ can be fed directly into subsequent layers of the network. In addition, the IC branch only introduces little computational cost, because its operations use lightweight structures.

We construct a set of IC networks by integrating the IC layers into existing models. Our experiments show that the IC networks have significant improvements compared to basic models with little additional computational burden. However, training from scratch may take a lot of time and suffer from complex hyper-parameter settings, especially when the basic models are large. From~\cite{DBLP:journals/corr/KirkpatrickPRVD16}, we find that the basic models may have similar parameter configurations to the corresponding IC models. Therefore, we propose a training method, namely the weak logit distillation (WLD), which distills the knowledge of pre-trained basic models. By combining the optimal basic models with a loss using weak soft targets, we show that the WLD only needs a few training rounds to successfully achieve or even exceed the result of training from scratch. In summary, our contributions are:
\begin{itemize}
    \item We propose a novel structure called the IC layer that can be used to build CNN architectures. We prove its effectiveness on enhancing representation and integrate it into some existing models. The experiments show the universality and superiority of the IC layer.
    \item We propose a method called WLD that can guide the learning of IC networks. It is a novel idea whose goal is to extract knowledge from smaller teacher models through the knowledge distillation (KD) method. The experiments show that by the WLD, the IC networks achieve higher performance with a shorter training time. Especially, the IC-ResNet-50 integrates the IC layer into ResNet-50 and reduces the top-1 error from $22.38\%$ to $21.75\%$, which also achieves the top-1 error of ResNet-100 ($21.75\%$).
\end{itemize}
% \item We propose a novel structure called the IC layer that can improve the performance of CNNs. We show that some existing models can achieve higher accuracy with the IC layers on the ImageNet dataset. The experiments show the universality and superiority of the IC layers.
    % \item We propose a training method called WLD that can train IC networks quickly while achieving the accuracy of training from scratch. 

% model-1.pdf

\section{Related Work}
\textbf{Effective computational unit.} Designing effective basic CNN units was a significant topic, since it reduces the difficulty of designing architectures by directly using basic units in existing models. \cite{hu2018squeeze} proposed the SE block that used feature recalibration to improve the performance of existing networks. However, the SE blocks were usually combined with building blocks rather than more basic structures. \cite{wang2019kervolutional} introduced the non-linear kernel method to convolutional layers to improve representation. Although that work bypassed the explicit calculation of the high dimensional features via a kernel trick, the complexity is obviously increased. In contrast to these work. Our proposed IC branch introduces the non-linear representation by a lightweight structure. Besides, it is combined with a single convolutional layer that can be applied to a wider range of CNN architectures. 

\textbf{Knowledge Distillation.} Recent KD methods used the feature distillation~\cite{DBLP:conf/iccv/HeoKYPK019} or self-supervision~\cite{DBLP:conf/eccv/XuLLL20} to extract the deep knowledge from a larger teacher model. Different from them, our training method WLD distill knowledge from a smaller pre-trained model while tolerating the gap between the teacher and student models. WLD is novel because it used the KD method to solve a different task: learning the optimal representation quickly and precisely when new components are introduced into the CNNs. The loss of WLD refers to the common KD loss proposed in~\cite{DBLP:journals/corr/HintonVD15}.

\section{The Inter-layer Collision Network}
In this section, we first show how the IC structure works and its combination with existing CNNs. Then, we introduce the WLD method to optimize IC networks. Finally, we analyze the influence of IC structure on model complexity.
\subsection{Optimization of non-linear representation of the MP model}\label{sec3_1}
The MP neuron~\cite{mcculloch1990logical} is the most commonly used neuron model, which can be formulated as $\mathrm{f}(\sum_{i=1}^{n} \ w_{i}x_{i} + b)$, where a linear transformation and a non-linear activation function $\mathrm{f}(\cdot)$ are applied to the $n$-D input successively. $w_{i}$ and $b$ denote the learnable weight and the bias, respectively. To facilitate the non-linear representation of neural models, we propose a new neuron model by replacing the linear transformation with a non-linear one: 
\begin{equation}
\label{eq1}
\begin{aligned}
\centering
y &=  \mathrm{f} \left( \sum_{i=1}^{n} \ w_{i}x_{i} + \sigma \left(\sum_{i=1}^{n} \ (w_{i} - 1)x_{i} + b_{1} \right) + b_{2} \right) \\
&= \mathrm{f} \left( \sum_{i=1}^{n} \ w_{i}x_{i} + \sigma \left(\sum_{i=1}^{n} \ w_{i}x_{i} - x_{sum} + b_{1} \right) + b_{2} \right) ,\\
\end{aligned}
\end{equation}
where $\mathrm{f}(\cdot)$ denotes a non-linear activation function and $\sigma(\cdot)$ is a rectified linear unit (ReLU)~\cite{nair2010rectified}. $b_{1}$ and $b_{2}$ are two independent biases used to adjust the center of the model distribution. $x_{sum}=\sum_{i=1}^{n} \ x_{i}$ denotes the summation of all the input features. We term the structure defined in eq.~\eqref{eq1} the \emph{inter-layer collision} (IC) neuron, since this idea is inspired by the physical elastic collision model where the speeds of two objects are $\frac{2m_{1}}{m_{1}+m_{2}}v$ and $(\frac{2m_{1}}{m_{1}+m_{2}}-1)v$ after collision (Details in Appendix C). We treat $\frac{2m_{1}}{m_{1}+m_{2}}$ as learnable weight $w_{i}$ and introduce some mathematical adjustments. 
% This structure enables the network to represent the non-linear distributions of the input signal~\cite{}. One significant advantage of the MP neuron is that the $\sum_{i=1}^{n} \ w_{i}x_{i}$ term which is continuous without any restrictions can flexibly represent any linear combinations in the input space. However, the representation ability of the MP neuron is simple so that real world tasks usually need large networks stacked by lots of neurons to fit complex distributions.

\begin{figure}
	\centering
	\subfigure[]{
		\centering
		\includegraphics[height=2.4cm, width=2.4cm]{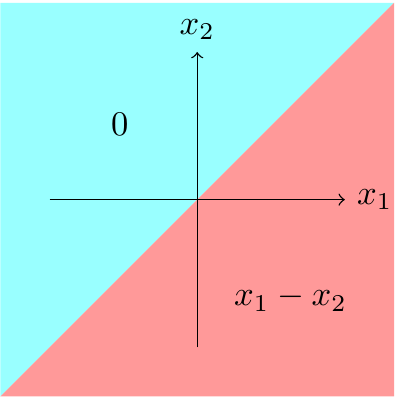}
	}
	\subfigure[]{
		\centering
		\includegraphics[height=2.4cm, width=2.4cm]{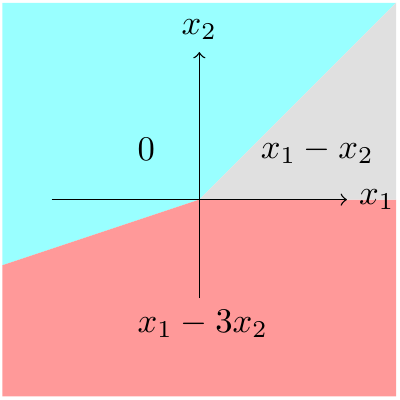}
	}
	\subfigure[]{
		\centering
		\includegraphics[height=2.4cm, width=2.4cm]{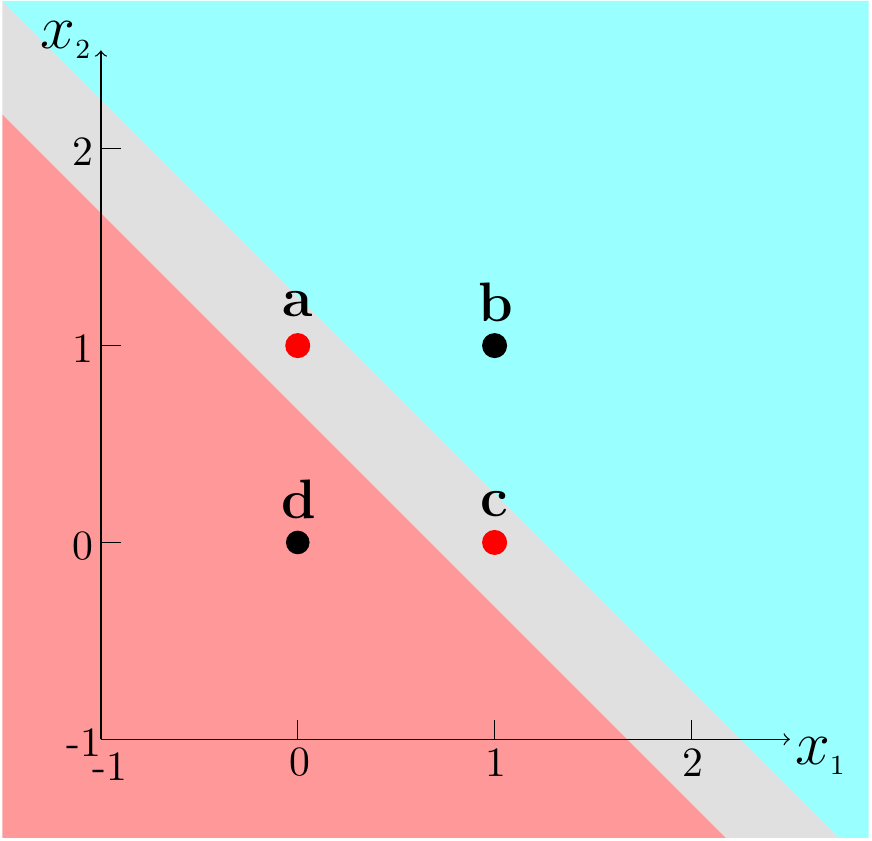}
	}
	\caption{The value of $f(x_{1},x_{2})$ given $x_{1}, x_{2}$. (a): $f(x_{1},x_{2})=\sigma (x_{1}-x_{2})$. (b): $f(x_{1},x_{2})=\sigma (x_{1}-x_{2}+\sigma(-2x_{2}))$. (c): $f(x_{1},x_{2})=\sigma (w_{1}x_{1}+w_{2}x_{2}+b_{1}+\sigma((w_{1}-1)x_{1}+(w_{2}-1)x_{2}+b_{2}))$. Here $w_{1} = w_{2} = 0.2805$. $b_{1} = -0.3506$ and $b_{2} = 0.6463$ are used to shift the boundary across the whole space. }
	\label{fig2}
\end{figure}

Through introducing the $\sigma(\cdot)$ operation, the neuron model can definitely increase the number of non-linear patterns. We use the term $H = \sum_{i=1}^{n} \ (w_{i} - 1)x_{i}$ to represent a hyperplane ($H=0$) in an $n$-dimensional Euclidean space, which divides eq.~\eqref{eq1} as follows:
\begin{equation}
\label{eq2}
y =
\begin{cases}
\mathrm{f} \left( 2\sum_{i=1}^{n} \ w_{i}x_{i} - \sum_{i=1}^{n} \ x_{i} \right) & \text{if } H \ge 0\\
\mathrm{f} \left(\sum_{i=1}^{n} \ w_{i}x_{i} \right) & \text{if } H < 0
\end{cases}.
\end{equation}
Here we omit the bias term. Intuitively, this IC neuron has a stronger representation ability than the MP neuron, since it can produce two different linear representations before the activation operation $\mathrm{f}(\cdot)$. To facilitate understanding, we use 2-D data as input and ReLU as the activation function to show how the two kinds of neurons generate non-linear boundaries. Figure~\ref{fig2}(a)(b) shows that the single MP neuron and the IC neuron can divide the $2$-D Euclidean space into multiple subspaces, each of which can represent a fixed linear transformation. We observe that a single IC neuron can divide one more subspace to represent a different linear transformation. Furthermore, we map the XOR problem which is a typical linear inseparable problem onto a $2$-D plane to explain the difference between the two kinds of neurons. For the ReLU MP neuron, it is clear that at least two neurons are required to solve the XOR problem. However, the non-linear boundary of the IC neuron is a broken line, providing a single neuron possibility to solve the XOR problem. Figure~\ref{fig2}(c) gives a solution with a single IC neuron, dividing the whole plane into three spaces where $(0,0),(1,1)$ are represented by zero and $(1,0),(0,1)$ are in two spaces with similar representation. 
% Figure \ref{fig3}(a) shows the calculation of a hyperplane in a $3$-D space where $\mathbf{W}=(w_{1}, w_{2}, w_{3})$ denotes the connection weights and $\mathbf{1}=(1, 1, 1)$.

Although neuron using the hyperplane $H=0$ can increase the number of non-linear patterns, the calculation of the hyperplane is limited by weight $w_{i}$, making it inflexible to divide different subspaces. Hence, the subspaces divided by the hyperplane are usually not optimal thus the weights easily converge to a local minimum. To add more representation flexibility to the hyperplane $H=0$, we improve eq.~\eqref{eq1} by introducing an adjustment weight $w'$:
\begin{equation}
\label{eq3}
\begin{aligned}
\centering
y &= \mathrm{f} \left( \sum_{i=1}^{n} \ w_{i}x_{i} + \sigma \left(\sum_{i=1}^{n} \ w_{i}x_{i} - w'x_{sum} + b_{1} \right) + b_{2} \right)\\
&= \mathrm{f} \left( \mathbf{w}^{T}\mathbf{x} + \sigma \left(\mathbf{w}^{T}\mathbf{x} - w' \times \mathbf{1}^{T}\mathbf{x} + b_{1} \right) + b_{2} \right) , \\
\end{aligned}
\end{equation}
where $\mathbf{w}$ and $\mathbf{x}$ represent weight vector and input vector, respectively. $\mathbf{1}$ is an all-one vector. $w'$ can be regarded as the intrinsic weight of one neuron, which is different from the weight $w_{i}$ connecting two neurons. Then there are two independent parameters in the calculation of the hyperplane $H=0$: $w'$ is used to change the direction of the hyperplane, and the bias $b_{1}$ is used to shift the hyperplane in the whole space. We give a theoretical analysis of the adjustable range of $w'$:
\newtheorem{theorem}{Theorem}
\begin{theorem}
	\label{th1}
	$\mathbf{w} = (w_{1},\dots,w_{n})^{T}$ and $\mathbf{1} = (1,\dots,1)^{T}$ are two n-$D$ vectors. By adjusting $w'$, the hyperplane $\sum_{i=1}^{n} \ (w_{i} - w')x_{i} = 0$ can be rotated $\pi$ $rad$ around the cross product of $\mathbf{w}$ and $\mathbf{1}$ when the two vectors are linearity independent.
\end{theorem}

Theorem~\ref{th1} implies that $\sum_{i=1}^{n} \ (w_{i} - w')x_{i} = 0$ can almost represent all the hyperplanes parallel to the cross product of $\mathbf{w}$ and $\mathbf{1}$, providing flexible strategies for dividing spaces. Besides, the IC neuron keeps a significant advantage over the MP neuron in that the $\sum_{i=1}^{n} \ w_{i}x_{i}$ term can flexibly represent any linear combinations in the input space. We provide the theorem as below:
% To further illustrate the superiority of the IC neuron, we provide the theorem \ref{th2} as below:
% \newtheorem{theorem}{}
\begin{theorem}
	\label{th2}
	In a closed $n$-D input space, for any given MP neuron $(w_{1}, \dots, w_{n}, b)$, there is always an IC neuron $(w_{1}, \dots, w_{n}, w', b_{1}, b_{2})$ that can completely represent this MP neuron.
\end{theorem}
The proof of Theorem~\ref{th1} and \ref{th2} are provided in the Appendix A. In summary, by adjusting the relationship between $\mathbf{w}$ and $w'$, the IC neuron can retain the representation ability of the MP neuron and flexibly segment linear representation spaces for some complex distributions. 
% We provide proof of theorem ~\ref{th1} and an example of $3$-D input space in the Appendix to facilitate understanding.
\subsection{Application on convolutional structure}\label{sec3_2}
% In a sliding window, the input feature also pass through a linear transformation and a non-linear operator.
The convolutional kernel, a filter used to capture the latent features in input signals, can be regarded as a combination of the MP model and a sliding window. To simplify the notation, we omit the activation operator and bias term. The output feature $\mathbf{u}_{i} \in \mathbb{R}^{H \times W}$ of the standard convolutional layer is given by: 
\begin{equation}
    \label{eq4}
	\begin{aligned}
		\mathbf{u}_{i} = \mathbf{w}_{i} \otimes \mathbf{X}, \\
	\end{aligned}
\end{equation}
where $\mathbf{X} \in \mathbb{R}^{H' \times W' \times C'}$ is the input feature map, $\mathbf{w}_{i} \in \mathbb{R}^{k \times k \times C'}$ is a filter kernel that belongs to $\mathbf{W}_{i} = [\mathbf{w}_{1}, \dots, \mathbf{w}_{C}]$ and $\otimes$ is used to denote the convolution operator. To apply the IC neuron model to eq.~\eqref{eq4}, we replace the kernel $\mathbf{w}_{i}$ with an IC kernel $[\mathbf{w}_{i}, w'_{i}]$, 
\begin{equation}
	\label{eq5}
	\begin{aligned}
		\mathbf{u}_{i} &= [\mathbf{w}_{i}, w'_{i}] \otimes \mathbf{X}\\
		&= \mathbf{w}_{i} \otimes X + \sigma(\mathbf{w}_{i} \otimes \mathbf{X} - w'_{i} \times (\mathbf{1} \otimes \mathbf{X})), \\ 
	\end{aligned}
\end{equation}
where $\mathbf{1}$ is an all-one tensor with the same size as $\mathbf{w}_{i}$. The input feature map may contain hundreds of channels and $\mathbf{1} \otimes \mathbf{X}$ term will mix features in all the channels with the same proportion. Therefore, we distinguish different features by a grouped convolutional trick:
\begin{equation}
	\label{eq6}
	\begin{aligned}
		\mathbf{u}_{i} = \mathbf{w}_{i} \otimes \mathbf{X} + \sigma(\mathbf{w}_{i} \otimes \mathbf{X} - (\mathbf{1} \hat{\otimes} \mathbf{X}) \cdot \mathbf{w}'_{i}), 
	\end{aligned}
\end{equation}
% The information $\mathbf{1}\hat{\otimes}\mathbf{X}$ contains some low-level features from previous layer, since it provides an approximate distribution of the pixels in the input feature maps, which helps the filters to learn high-level features faster. 
where $\hat{\otimes}$ denotes the depthwise convolution \cite{chollet2017xception} which separates $\mathbf{1}$ and $\mathbf{X}$ into $C'$ independent channels and performs channel-wise convolution. The adjustment weight $\mathbf{w}'_{i}$ becomes a vector with size $C'$, recalibrating the features of $\mathbf{1}\hat{\otimes}\mathbf{X}$ by channel-wise multiplication ($\cdot$). Note that all the convolution operators ($\otimes$ and $\hat{\otimes}$) in eq.~\eqref{eq6} share the same stride and padding. The structure in eq.~\eqref{eq6} which we term the IC layer has two main advantages compared to the traditional convolutional layer:
\begin{itemize}
    \item According to Section~\ref{sec3_1}, the single kernel $[\mathbf{w}_{i}, \mathbf{w}'_{i}]$ can represent more linear patterns before activation, which enhancing the representation of convolutional layer.
    \item The information $\mathbf{1}\hat{\otimes}\mathbf{X}$ contains some low-level features from the previous layer. It helps the filters to learn high-level features faster since it provides an approximate distribution of the pixels in the input feature maps. 
\end{itemize}
The IC layer can be easily integrated into some existing models. Consider the ResNets~\cite{he2016deep} which are commonly used networks as examples. The basic block used by ResNet-18 and ResNet-34 has two $3 \times 3 $ convolutional layers. The bottleneck block used by deeper ResNets has three convolutional layers ($1\times1, 3\times3, 1\times1$). Note that the information $\mathbf{1}\hat{\otimes}\mathbf{X}$ equals $\mathbf{X}$ when kernel size is $1\times1$, it will not provide extra features for the filter. Therefore, we prefer to replace all the $3\times3$ layers in building blocks to build IC-ResNets. The combination with other popular models is introduced in Section 4, where we show the universality and superiority of the IC layer.

% Besides, our experiments show that although the $1\times1$ IC layers can improve the performance of the networks, they aggravate overfitting and bring a non-negligible additional cost. Therefore, we prefer to replace only $3\times3$ layers to build IC-ResNets. The combination with other popular models is introduced in Section 4, where we show the universality and superiority of the IC layer.
% We consider that it is unnecessary to implement the IC structure on a $1 \times 1$ convolution, since the information $\mathbf{1}**\mathbf{X} = \mathbf{X}$ which make . 
% Moreover, the IC networks tend to capture the all target objects in the images!. Besides, the IC networks distinguish the target object more accurately.
\subsection{Learning with Knowledge Distillation}\label{sec3_3}
\begin{figure}
		\centering
	\subfigure{
		\centering
		\includegraphics[scale=0.186]{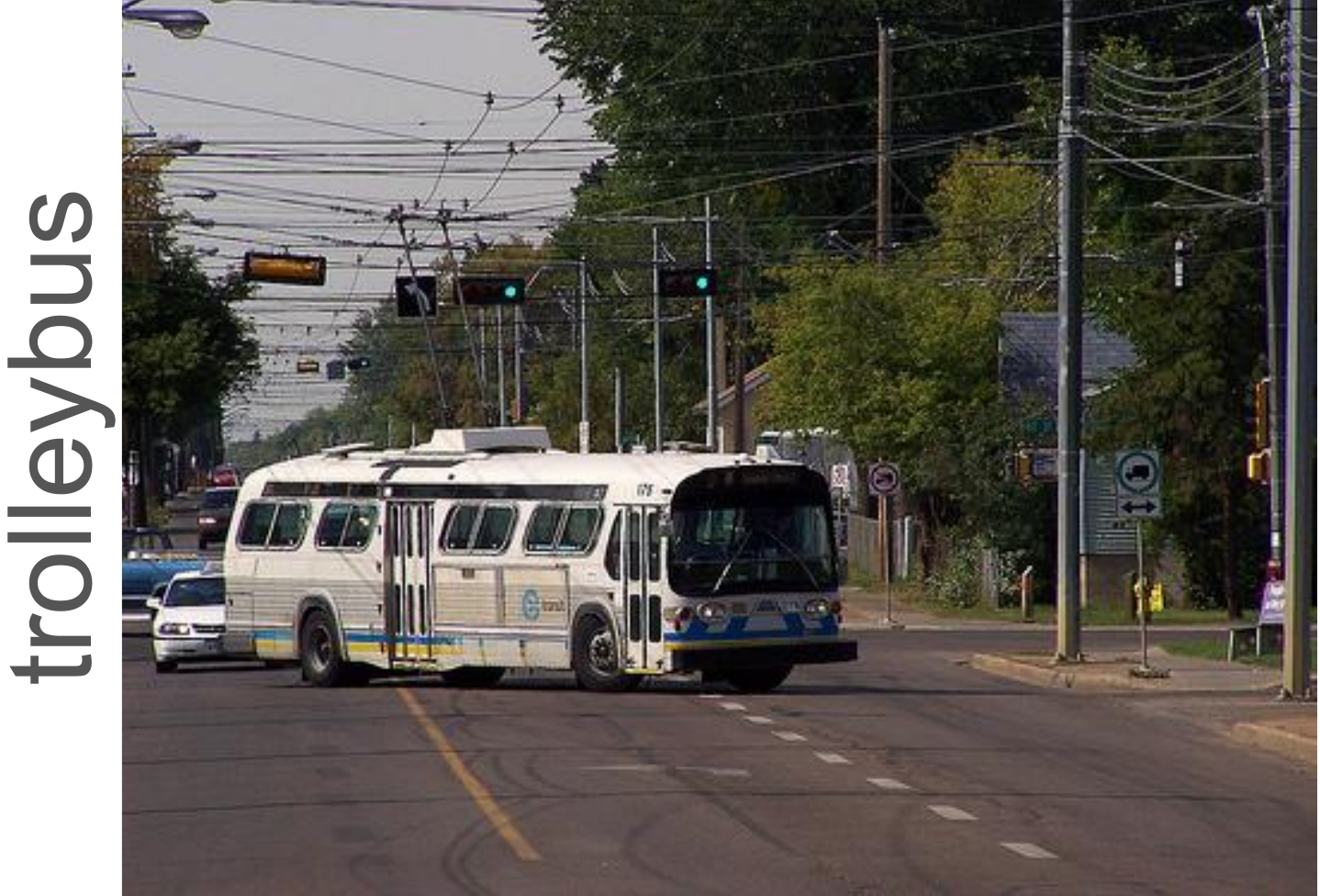}
	}
	\subfigure{
		\centering
		\includegraphics[scale=0.14]{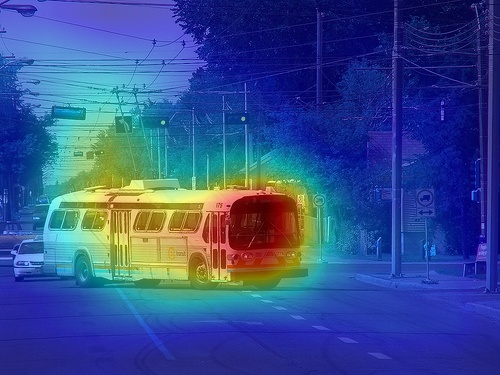}
	}
	\subfigure{
		\centering
		\includegraphics[scale=0.139]{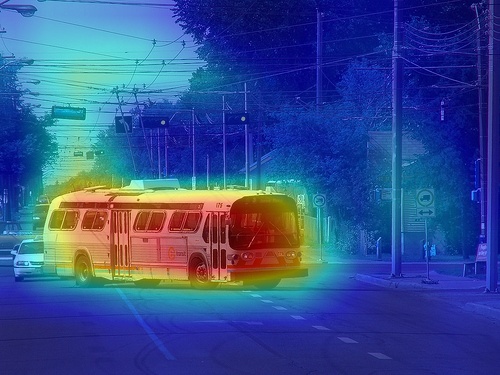}
	}
	\subfigure[original image]{
		\centering
		\includegraphics[scale=0.186]{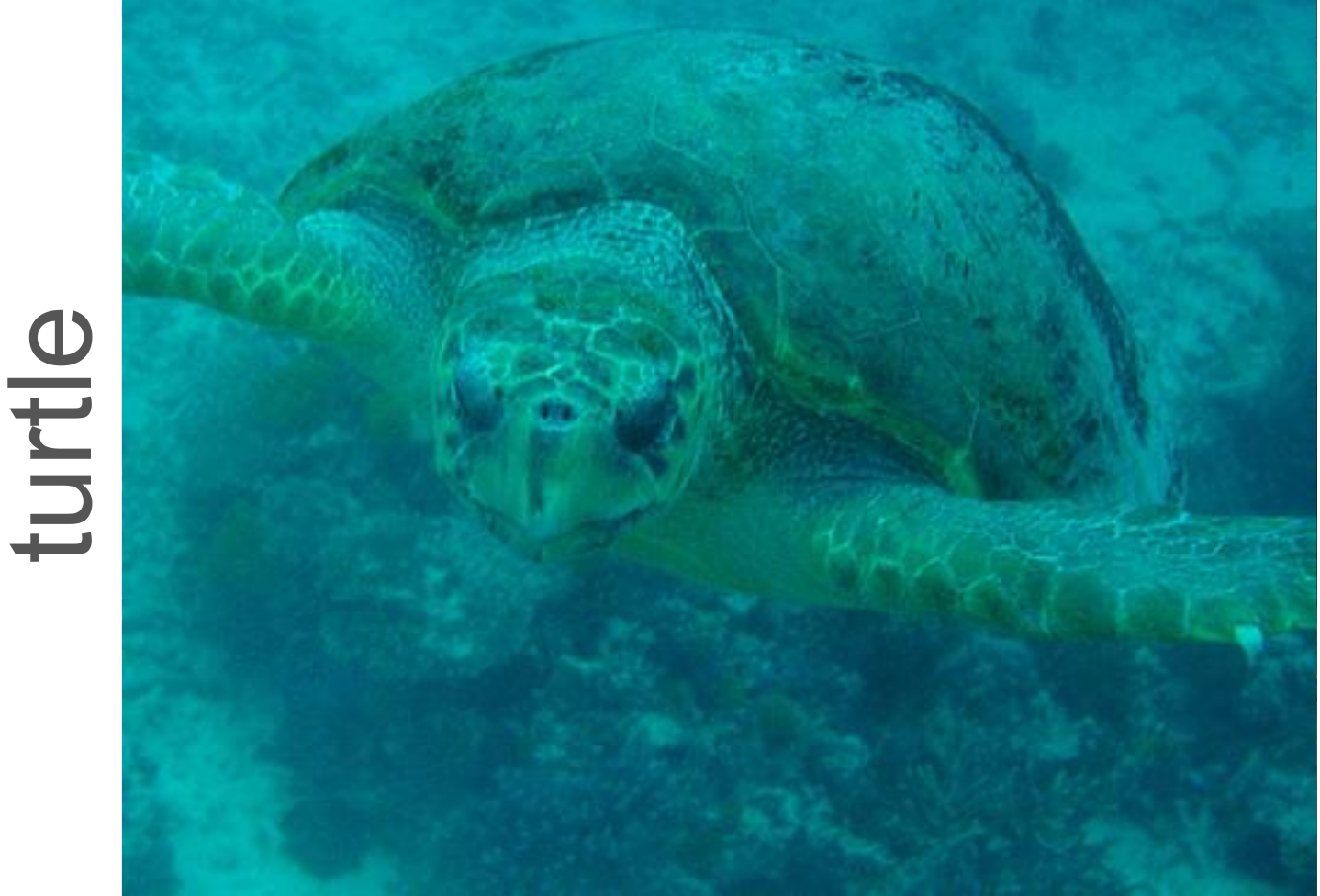}
	}
	\subfigure[ResNet-18]{
		\centering
		\includegraphics[scale=0.14]{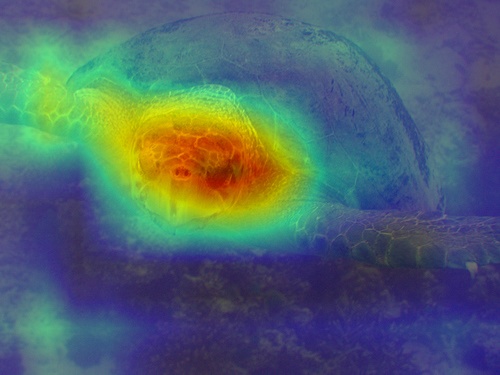}
	}
	\subfigure[IC-ResNet-18]{
		\centering
		\includegraphics[scale=0.14]{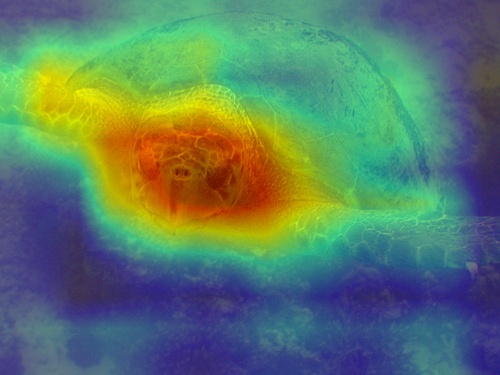}
	}
	% \hspace{1cm}
	% \centering 
	\caption{The Grad-CAM visualizations for different models. The images are randomly picked from ImageNet validation set.}
	\label{fig3}
	\end{figure}

In order to further understand why the IC layer can capture more fine-grained features, we compare the Grad-CAM visualizations~\cite{DBLP:conf/iccv/SelvarajuCDVPB17} of the IC models with their basic models. As shown in Figure~\ref{fig3}, the IC networks tend to focus on more relevant regions with more object details. More importantly, the features of IC networks and basic networks have some similarities. We think that although the features captured by IC networks have finer texture information, their focus on the image is similar to features captured by basic networks. 
% (The Table~\ref{} shows that this similarity is also observed on the final layer logits between IC networks and their basic networks. Appendix)
\makeatletter
\renewcommand{\@thesubfigure}{\hskip\subfiglabelskip}
\makeatother

Motivated by the observed similarity, we propose a method to guide the learning of IC networks by using the knowledge of pre-trained basic networks. First, we have a basic network \emph{B} with a pre-trained set of parameters $\mathbf{\theta}^{*}$ and our goal is to train a corresponding IC network \emph{IC-B} with better performance. According to Theorem~\ref{th2} and the argument that a network has many configurations of parameters with the same performance~\cite{DBLP:journals/corr/KirkpatrickPRVD16}, we assume that \emph{IC-B} can have a configuration close to $\mathbf{\theta}^{*}$. Based on our hypothesis, we load the set $\mathbf{\theta}_{*}$ for \emph{IC-B} and add a scaling factor $\alpha$ to control the influence of $\sigma(\cdot)$. The IC layer in \emph{IC-B} can be initialized as:
\begin{equation}
\label{eq7}
	\begin{aligned}
		\mathbf{u}_{i} = \mathbf{w}_{i}^{*} \otimes \mathbf{X} + \alpha \times \sigma(\mathbf{w}_{i}^{*}  \otimes \mathbf{X} - (\mathbf{1} \hat{\otimes} \mathbf{X}) \cdot \mathbf{w}'_{i}), 
	\end{aligned}
\end{equation}
% The $\sigma(\cdot)$ term can be regarded as adjusting a prior knowledge by the adjustment weight $\mathbf{w}'_{i}$
where $\mathbf{w}_{i}^{*}$ is a weight loaded from $\mathbf{\theta}_{*}$ and we use random initialization for $\mathbf{w}'_{i}$. Since the $\mathbf{w}_{i}^{*} \otimes \mathbf{X}$ term can capture the features independently, $\alpha$ is set to zero at the beginning to let \emph{IC-B} have the same performance as \emph{B}. After initialization, we fine-tune all the parameters. The $\sigma(\cdot)$ term benefits from the pre-trained information, making the adjustment weight converges quickly. We use the scaling factor $\alpha$ to gradually amplify the influence of $\sigma(\cdot)$ in the training process. There are two strategies to adjust $\alpha$. The first one is to gradually increase its value manually, but it need to provide accurate hyperparameters. The second is to treat $\alpha$ as a parameter trained with other parameters. Our experience shows that the best value of $\alpha$ is usually in $[0.01, 0.04]$. 

To further use the knowledge of \emph{B}, we refer to the way in KD. We use the basic network \emph{B} as a teacher model and the \emph{IC-B} as a student model. The soft targets predicted by a well-optimized teacher model can provide extra information, compared to the ground truth. To obtain the soft targets of \emph{B}, temperature scaling~\cite{DBLP:journals/corr/HintonVD15} is used to soften the peaky softmax distribution:
\begin{equation}
\label{eq8}
	\begin{aligned}
		p^{i}(x;\tau)=\frac{e^{s_{i}(x)/\tau}}{\sum_{k} e^{s_{k}(x)/\tau}}, 
	\end{aligned}
\end{equation}
where $x$ is the data sample, $i$ is the category index, $s_{i}(x)$ is the score logit that $x$ obtains on category $i$, and $\tau$ is the temperature. The knowledge distillation loss $L_{kd}$ measured by KL-divergence as:
\begin{equation}
\label{eq9}
	\begin{aligned}
		L_{kd}=-\tau^{2}\sum_{x\in \mathcal{D}_{x}} \sum_{i=1}^{C} p_{t}^{i}(x;\tau) \log (p_{s}^{i}(x;\tau)),
	\end{aligned}
\end{equation}
where $t$ and $s$ denote the teacher and student models. $C$ is the total number of classes, $\mathcal{D}_{x}$ indicates the dataset. The complete loss function $L$ of training \emph{IC-B} is a combination of the standard cross-entropy loss $L_{ce}$ and knowledge distillation loss $L_{kd}$:
\begin{equation}
\label{eq10}
	\begin{aligned}
		L= (1-\lambda) L_{ce} + \lambda \max (L_{ce} - e, 0),
	\end{aligned}
\end{equation}
where $\lambda$ is a balancing weight. $e$ is a constant used to increase tolerance the for the gap between the soft targets of \emph{B} and \emph{IC-B}.
% (Usually use $\lambda_{1}:0.9,\lambda_{2}:0.1$)

% \begin{table*}
% 	\centering
% 	\setlength{\tabcolsep}{7mm}{
% 	\begin{tabular}{l|cc|cc|cc}  
% 	\toprule
% 	Model  & \multicolumn{2}{|c|}{Top-1 err.} & \multicolumn{2}{|c|}{Top-5 err.} & GFlops & Params \\
% 	\midrule
% 	ResNet-18       & 30.24 & 27.88 & 10.92 & 9.42 & 1.82 & 11.7M\\
% 	IC-ResNet-18	& 28.56 & 26.69 & 9.80 & 8.56 & 2.01  & 12.9M\\
% 	\hline
% 	ResNet-34		& 26.70 & 24.52 & 8.58 & 7.46 & 3.68 & 21.8M\\
% 	IC-ResNet-34	& 25.55 & 23.49 & 7.90 & 6.86 & 4.07 & 24.2M\\
% 	\hline  
% 	ResNet-50		& 23.85 & 22.85 & 7.13 & 6.71 & 4.12 & 25.6M\\
% 	IC-ResNet-50	& 23.20 & 21.90 & 6.72 & 6.08 & 4.33 & 26.8M\\
% 	\bottomrule
% 	\end{tabular}
% 	}
% 	\caption{IC-ResNets performance results on ImageNet validation set. The error rate use }
% 	\label{tab1}
% \end{table*}
\begin{table*}
	\centering
	\setlength{\tabcolsep}{8.5mm}{
	\begin{tabular}{l|cc|cc}  
	\toprule
	Model  & Top-1 err. & Top-5 err. & GFlops & Params \\
	\midrule
	ResNet-18       & 30.24 / 27.88 & 10.92 / 9.42 & 1.82 & 11.7M \\
	IC-ResNet-18	& \textbf{28.56} / \textbf{26.69} & \textbf{9.80} / \textbf{8.56} & 2.01  & 12.9M \\
	\hline
	ResNet-34		& 26.70 / 24.52 & 8.58 / 7.46 & 3.68 & 21.8M \\
	IC-ResNet-34	& \textbf{25.55} / \textbf{23.49} & \textbf{7.90} / \textbf{6.86} & 4.07 & 24.2M \\
	\hline  
	ResNet-50		& 23.85 / 22.85 & 7.13 / 6.71 & 4.12 & 25.6M \\
	IC-ResNet-50	& \textbf{23.20} / \textbf{21.90} & \textbf{6.72} / \textbf{6.08} & 4.33 & 26.8M \\
	\bottomrule
	\end{tabular}
	}
	\caption{IC-ResNets performance results on ImageNet validation set. The error rates (\%) use single-crop/10-crop testing. }
	\label{tab1}
\end{table*}

Different from the traditional KD method, our method does not extract the knowledge from a deeper network with better performance. Therefore, we do not need the output distributions of the teacher and student models to be exactly equal. The loss term $\max(L_{ce} - e, 0)$ allows deviation between $p_{s}^{i}$ and $p_{t}^{i}$. Besides, during training, we gradually reduce the value of $\lambda$ to make the \emph{IC-B} reduce the dependence on the teacher network. Eq.~\eqref{eq10} provides a strategy to introduce some extra information for training IC networks. Combined with loading pre-trained parameters and fine-tuning, the IC networks achieve high performance after a few learning rounds. We term this learning method the \emph{weak logit distillation} (WLD) since it weakens the impact of soft targets to reduce the dependence on the teacher networks.

\subsection{Parameters and Complexity Analysis}\label{sec3_4}
For the standard convolutional layer with $k \times k$ receptive fields, the transformation
\begin{equation}
	\begin{aligned}
		\mathbf{X} \in \mathbb{R}^{H' \times W' \times C'} \xrightarrow{conv} \mathbf{U} \in \mathbb{R}^{H \times W \times C}\\
	\end{aligned}
\end{equation}
needs $k \times k \times C' \times C$ parameters. In the IC layer, we calculate the sum of each channel without additional parameters. The weight $\mathbf{W'} = [\mathbf{w'}_{1},\mathbf{w'}_{2},\cdots,\mathbf{w'}_{C}]$ adds $1 \times 1 \times C' \times C$ parameters. Therefore, the number of parameters added by the IC layer is only $\frac{1}{k \times k}$ of the original layer.

The IC layer adds a depthwise convolution and a learnable weight $\mathbf{W}'$ which can be regarded as a $1 \times 1$ convolution. The depthwise convolution is used to calculate the element-wise sum of the input by an all-one kernel. The increased computational complexity is the same as adding a convolutional filter because we only need to do this operation once. Therefore, the increased complexity is $\frac{1}{C}$ of the original layer. The weight $\mathbf{W'}$ is a $1 \times 1$ convolution which uses less than $\frac{1}{k \times k}$ computation of the $k \times k$ convolution \cite{sifre2014rigid}. Therefore, the approximate extra computation is $\frac{1}{C} + \frac{1}{k \times k}$ of the original layer. 

\section{Experiments}
In this section, we investigate the effectiveness of different IC architectures by a series of comparative experiments. Besides, we evaluate the WLD method on training IC networks. 

% Besides, our experiments show that although the $1\times1$ IC layers can improve the performance of the networks, they aggravate overfitting and bring a non-negligible additional cost. Therefore, we prefer to replace only $3\times3$ layers to build IC-ResNets. The combination with other popular models is introduced in Section 4, where we show the universality and superiority of the IC layer.

\subsection{ImageNet Results}
We use the ILSVRC 2012 classification dataset \cite{russakovsky2015imagenet} which consists of more than 1 million color images in 1000 classes divided into 1.28M training images and 50K validation images. We use three versions of ResNets (ResNet-18, ResNet-34, ResNet-50) to build the corresponding IC networks. For a fair comparison, all our experiments on ImageNet are conducted with the same environment setting. The optimizer uses the stochastic gradient descent (SGD) \cite{lecun1989backpropagation} method with a weight decay of $10^{-4}$ and a momentum of $0.9$. The training process is set to $120$ epochs with a batch size of $256$. The learning rate is initially set to $0.1$ and will be reduced 10 times every $30$ epochs. Besides, all experiments are implemented with the Pytorch~\cite{paszke2019pytorch} framework on a server with NVIDIA TITAN Xp GPUs.

\begin{table}\Large
	\centering
	\normalsize
	\setlength{\tabcolsep}{5mm}{
	\normalsize
	\begin{tabular}{l|cc}  
% 	\small
	\toprule
	\normalsize
	Model  & Top-1 err. &  Top-5 err. \\
	\midrule
	IC-ResNet-18 & 28.82 & 9.98  \\
	IC-ResNet-34 & 25.78 & 8.02   \\
	IC-ResNet-50 & 23.07 & 6.67   \\
	\hline
	EfficientNet-B0 & 23.7 & 6.8      \\
	IC-EfficientNet-B0 & \textbf{23.6} & \textbf{6.7} \\
	\hline
	EfficientNet-B1 & 21.2 & 5.6      \\
	IC-EfficientNet-B1 & \textbf{21.0} & \textbf{5.5}  \\
	\hline
	EfficientNet-B2 & 20.2 & 5.1      \\
	IC-EfficientNet-B2 & \textbf{20.1} & \textbf{5.0}  \\
	\bottomrule
	\end{tabular}
	}
	\caption{The single-crop error rates (\%) of WLD results on the ImageNet validation set.}
	\label{tab2}
\end{table}

% \begin{figure*}[h]
% 	\label{Imagenet}
% 	\centering
% 	\subfigure[]{
% 	\centering
% 	% \newcommand{scale1}{0.3}
% 	\includegraphics[scale = 0.4]{vgg16-compare-crop2.pdf}
% 	}
% 	% \hspace{0.5cm}
% 	\subfigure[]{
% 	\centering
% 	\includegraphics[scale = 0.4]{vgg19-compare-crop2.pdf}
% 	}
% 	% \hspace{0.5cm}
% 	\subfigure[]{
% 	\centering
% 	\includegraphics[scale = 0.4]{mobile-compare-crop2.pdf}
% 	}
% 	\subfigure[]{
% 	\centering
% 	\includegraphics[scale = 0.4]{senet18-compare-crop2.pdf}
% 	}
% 	% \hspace{0.5cm}
% 	\subfigure[]{
% 	\centering
% 	\includegraphics[scale = 0.4]{senet50-compare-crop2.pdf}
% 	}
% 	% \hspace{0.5cm}
% 	\subfigure[]{
% 	\centering
% 	\includegraphics[scale = 0.4]{ResNext-compare-crop2.pdf}
% 	}
% 	\caption{Training curves of the six IC networks and their basic models on CIFAR10. The IC networks use the IC layer in (a), (b), (c), and use the IC block in (d), (e), (f).} 
% 	\label{fig4}
% \end{figure*}
% We use the WLD method with $\lambda_{1}$ of $0.9$, $\lambda_{2}$ of $0.1$ and $e$ of $0.005$ to train IC-networks.
To compare with previous research~\cite{he2016deep,wang2019kervolutional}, we apply both the single-crop and 10-crop for testing. We build the IC-ResNet-18, IC-ResNet-34 and IC-ResNet-50 by replacing the whole $3 \times 3$ convolutional layers in blocks with $3 \times 3$ IC layers. The validation error and FLOPs are reported in Table \ref{tab1}. We observe that the IC-ResNet-18 and IC-ResNet-34 can obviously reduce the 10-crop top-1 error by $1.19\%, 1.03\%$ and the top-5 error by $0.86\%, 0.60\%$ with a small increase in the calculation ($10.44\%, 10.60\%$), validating the effectiveness of the IC layer. The IC-ResNet-50 retains $1\times1$ convolutional layers in bottleneck blocks. The 10-crop top-1 error is $21.90\%$ and the top-5 error is $6.08\%$, exceeding ResNet-50 by $0.95\%$ and $0.63\%$, respectively. Moreover, the extra FLOPs of the IC-ResNet-50 is only $5.10\%$ of the ResNet-50. For the deeper ResNets, we believe that the deeper IC-ResNets can get similar results, since they all use the same block as ResNet-50.

\subsubsection{Training with WLD}
To evaluate the WLD method, We construct a set of experiments to train the IC-ResNets and three versions of the EfficientNets~\cite{DBLP:conf/icml/TanL19} (EfficientNet-b0, EfficientNet-b1 and EfficientNet-b2). The training process is set to $30$ epochs and the learning rate whose initial value is $0.001$ will be reduced 10 times every $15$ epochs ($10$ epochs for EfficientNets). The other hyperparameter settings are: $e$ is $0.005$; $\lambda$ initially set to $0.9$ is reduced to $0.1$. Table~\ref{tab2} shows that WLD achieves the accuracy of training from scratch with fewer training rounds. Remarkably, by training with WLD, the 10-crop result of the IC-ResNet-50 ($21.75\%$ top-1 and $5.92\%$ top-5 error) achieves the error rate achieved by the deeper ResNet-101 network ($21.75\%$ top-1 and $6.05\%$ top-5 error) with nearly half of the computational burden ($4.33$ GFLOPs vs. $7.85$ GFLOPs). 

This set of experiments prove our hypothesis mentioned in Section~\ref{sec3_3} and provide a understandable conclusion: although the low learning rate limits the connection weights in the vicinity of pre-trained model, the WLD method can still find optimal adjustment weights to improve the representation of IC networks. Specially, the EfficientNets which come from neural architecture search have difficuly in training from scratch. Through the pre-trained models and WLD method, the corresponding IC-EfficientNets achieve higher performance with a simple hyperparameter configuration. 

% Besides, when pre-trained models are not available, training from scratch can also make IC networks achieve higher performance. 

\subsubsection{The effectiveness of $1\times1$ IC layers}
To evaluate the $1\times1$ IC layers, we integrate them into the IC-ResNet-50 to build the IC-ResNet-50-B by only replacing the first $1\times1$ layers in bottleneck blocks and the IC-ResNet-50-C by replacing all $1\times1$ layers. As shown in Table~\ref{tab3}, although the IC-ResNet-50-B and IC-ResNet-50-C exceed the IC-ResNet-50, it obviously increases the FLOPs of the model. Besides, we observe that the overfitting in two IC models where the training accuracy is improved significantly but the test accuracy is not. Combined with analysis in Section~\ref{sec3_2}, we argue that $1 \times 1$ IC layers focus more on improving model capacity rather than introducing new feature information. When the number of channels of $1\times1$ IC layers is relatively large, the models are more likely to cause overfitting and bring an expensive computational burden. 

\begin{table}
	\centering
	\setlength{\tabcolsep}{4.1mm}{
	\begin{tabular}{l|cc}  
	\toprule
	Model  & Top-1 err. & GFlops/Params \\
	\midrule
	IC-ResNet-50 & 23.07/6.68 & 4.33/26.8M \\
	IC-ResNet-50-B & 23.02/6.66 & 5.27/31.2M \\ 
	IC-ResNet-50-C & 22.96/6.65 & 6.10/36.2M \\
	\bottomrule
	\end{tabular}
	}
	\caption{Performance results of the IC networks with $1\times1$ IC layers on ImageNet validation set. }
	\label{tab3}
\end{table}

\subsection{CIFAR-10 Results}
We further investigate the universality of the IC layer by integrating it into some other modern architectures. These experiments are conducted on the CIFAR10 dataset, which consists of 60K $32 \times 32$ colour images in 10 classes divided into 50K training images and 10K testing images. Each model is trained in $200$ epochs with a batch size of $128$. The learning rate is initialized to $0.1$, which will be reduced $10$ times at the $60$th epoch and the $120$th epoch. The optimizer settings are the same as in the ImageNet experiment.

% First, we use the first strategy mentioned in Section. 3.3 for VGGNet\cite{} and MobileNet\cite{} which have no block structures. We select two versions of VGGNet(VGG-16, VGG-19). Specially, we treat the two adjacent layers in MobileNet(a depthwise separable and a pointwise convolution) as one convolutional layer to add the IC structure, since there is a close relationship between adjacent layers. As shown in Fig. 7(a, b, c) and Table. 4, the networks with the IC structure can have greater performance on both training speed and error rate. 

We integrate the IC layers into VGGNets (VGG-16 and VGG-19 versions), MobileNet~\cite{howard2017mobilenets}, SENets (SE-ResNet-18 and SE-ResNet-50 versions) and ResNeXt~\cite{xie2017aggregated} (2x64d version). Specially, the adjacent convolution layers (a depthwise convolution layer and a pointwise convolutional layer) in MobileNet are treated as one convolutional layer when integrating with the IC layer because there is a close relationship between adjacent layers. The results are listed in Table \ref{tab4}, we observe that the IC layers improve representation of all the basic models. This set of experiments show the universality of the IC layers. Besides, we observe that the IC networks have faster convergence speed than basic models in both the ImageNet and CIFAR-10 experiments. The training curves are shown in Appendix B.

\begin{table}
	\centering
	\setlength{\tabcolsep}{1mm}{
	\begin{tabular}{lc|lc}  
	\toprule
	Model  & Top-1 err. & Model & Top-1 err. \\
	\midrule
	VGG-16      & 6.36  & SE-ResNet-18 & 5.08\\
	IC-VGG-16	& \textbf{6.04}  & IC-SE-ResNet-18 & \textbf{4.85}\\
	\hline
	VGG-19	& 6.46      & SE-ResNet-50 & 5.10\\
	IC-VGG-19	& \textbf{6.27}  & IC-SE-ResNet-50 & \textbf{4.61}\\
	\hline
	MobileNet & 9.92    & ResNext(2x64d) & 4.62\\
	IC-MobileNet & \textbf{9.00} & IC-ResNext(2x64d) & \textbf{4.43}\\
	\bottomrule
	\end{tabular}
	}
	\caption{Results on CIFAR10 with various IC models. We use our environment settings to reprodurce the baseline results.}
	\label{tab4}
	% \label{tab:booktabs}
\end{table}

\begin{table}
	\centering
 	\setlength{\tabcolsep}{6mm}{
	\begin{tabular}{l|cc}  
	\toprule
	framework & backbone  & mAP \\
	\midrule
    \multirow{2}*{Faster R-CNN} & ResNet-50 & 79.5\\
    ~ & IC-ResNet-50 & \textbf{80.5}\\
    \hline 
    \multirow{2}*{Retinanet} & ResNet-50 & 77.3\\
    ~ & IC-ResNet-50 & \textbf{78.2}\\
	\bottomrule
	\end{tabular}
	}
	\caption{Results on PASCAL VOC 2007+2012 test set. }
	\label{tab5}
	% \label{tab:booktabs}
\end{table}

\subsection{Object Detection}
We further assess the generalization of IC networks on the task of object detection using the PASCAL VOC 2007+2012 detection benchmark~\cite{DBLP:journals/ijcv/EveringhamGWWZ10}, This dataset consists of about 5K train/val images and 5K test images over 20 object categories. We use the IC-ResNet-50 as the backbone networks to capture the features. Weights are initialized by the parameters of the IC-ResNet-50 trained by WLD on the ImageNet dataset. The detection frameworks that we use are Faster R-CNN~\cite{DBLP:conf/nips/RenHGS15} and Retinanet~\cite{DBLP:conf/iccv/LinGGHD17}. We use the same configuration for both the IC-ResNet-50 and the ResNet-50, which is described in~\cite{DBLP:journals/corr/abs-1906-07155}. We evaluate detection mean Average Precision (mAP) which is the actual metric for object detection. As shown in Table~\ref{tab5}, IC-ResNet-50 outperforms ResNet-50 by $1.0\%$ and $0.9\%$ on the Faster R-CNN and Retinanet frameworks, respectively. The ResNet-50 result we report is the same as previous work. Our experiments demonstrate that the IC networks can be easily integrated into the object detection and achieve better performance with negligible additional cost. We believe that IC networks can show their superiority across a broad range of vision tasks and datasets.

\section{Conclusion}

In this paper, we propose the IC structure that brings non-linearity and feature recalibration to convolution operation. By dividing the input space, the IC structure has a stronger representation ability than the traditional convolution structure. We build the IC networks by integrating the IC structure into the state-of-the-art models. Besides, we propose the WLD method to facilitate the training of IC networks. It is shown that training with WLD can bypass requirement of the complex hyperparameter design and reach convergence quickly. A wide range of experiments show the effectiveness of IC networks across multiple datasets and tasks. Finally, we expect to integrate IC structure into more architectures and further improve the performance of computer vision tasks.

\newpage

\bibliographystyle{named}
\bibliography{ijcai21}

\begin{thebibliography}{}

\bibitem[\protect\citeauthoryear{Bell \bgroup \em et al.\egroup
  }{2016}]{bell2016inside}
S.~Bell, C.~L. Zitnick, K.~Bala, and R.~B. Girshick.
\newblock Inside-outside net: Detecting objects in context with skip pooling
  and recurrent neural networks.
\newblock In {\em CVPR}, pages 2874--2883, 2016.

\bibitem[\protect\citeauthoryear{Chen \bgroup \em et al.\egroup
  }{2019}]{DBLP:journals/corr/abs-1906-07155}
K.~Chen, J.~Q. Wang, J.~M. Pang, Y.~H. Cao, Y.~Xiong, X.~X. Li, S.~Y. Sun,
  W.~S. Feng, Z.~W. Liu, J.~R. Xu, et~al.
\newblock Mmdetection: Open mmlab detection toolbox and benchmark.
\newblock {\em arXiv:1906.07155}, 2019.

\bibitem[\protect\citeauthoryear{Chollet}{2017}]{chollet2017xception}
F.~Chollet.
\newblock Xception: Deep learning with depthwise separable convolutions.
\newblock In {\em CVPR}, pages 1251--1258, 2017.

\bibitem[\protect\citeauthoryear{Everingham \bgroup \em et al.\egroup
  }{2010}]{DBLP:journals/ijcv/EveringhamGWWZ10}
M.~Everingham, L.~Van Gool, C.~K.~I. Williams, J.~M. Winn, and A.~Zisserman.
\newblock The pascal visual object classes {(VOC)} challenge.
\newblock {\em International journal of computer vision}, 88(2):303--338, 2010.

\bibitem[\protect\citeauthoryear{He \bgroup \em et al.\egroup
  }{2016}]{he2016deep}
K.~M. He, X.~Y. Zhang, S.~Q. Ren, and J.~Sun.
\newblock Deep residual learning for image recognition.
\newblock In {\em CVPR}, pages 770--778, 2016.

\bibitem[\protect\citeauthoryear{Heo \bgroup \em et al.\egroup
  }{2019}]{DBLP:conf/iccv/HeoKYPK019}
B.~Heo, J.~Kim, S.~Yun, H.~Park, N.~Kwak, and J.~Y. Choi.
\newblock A comprehensive overhaul of feature distillation.
\newblock In {\em ICCV}, pages 1921--1930, 2019.

\bibitem[\protect\citeauthoryear{Hinton \bgroup \em et al.\egroup
  }{2015}]{DBLP:journals/corr/HintonVD15}
G.~E. Hinton, O.~Vinyals, and J.~Dean.
\newblock Distilling the knowledge in a neural network.
\newblock In {\em NeurIPS Deep Learning and Representation Learning Workshop},
  2015.

\bibitem[\protect\citeauthoryear{Howard \bgroup \em et al.\egroup
  }{2017}]{howard2017mobilenets}
A.~G. Howard, M.~l. Zhu, B.~Chen, D.~Kalenichenko, W.~J. Wang, T.~Weyand,
  M.~Andreetto, and H.~Adam.
\newblock Mobilenets: Efficient convolutional neural networks for mobile vision
  applications.
\newblock In {\em CVPR}, 2017.

\bibitem[\protect\citeauthoryear{Hu \bgroup \em et al.\egroup
  }{2018}]{hu2018squeeze}
J.~Hu, L.~Shen, and G.~Sun.
\newblock Squeeze-and-excitation networks.
\newblock In {\em CVPR}, pages 7132--7141, 2018.

\bibitem[\protect\citeauthoryear{Kirkpatrick \bgroup \em et al.\egroup
  }{2017}]{DBLP:journals/corr/KirkpatrickPRVD16}
J.~Kirkpatrick, R.~Pascanu, N.~Rabinowitz, J.~Veness, G.~Desjardins, A.~A.
  Rusu, K.~Milan, J.~Quan, T.~Ramalho, A.~Grabska-Barwinska, D.~Hassabis,
  C.~Clopath, D.~Kumaran, and R.~Hadsell.
\newblock Overcoming catastrophic forgetting in neural networks.
\newblock {\em Proceedings of the National Academy of Sciences},
  114(13):3521--3526, 2017.

\bibitem[\protect\citeauthoryear{Krizhevsky \bgroup \em et al.\egroup
  }{2012}]{krizhevsky2012imagenet}
A.~Krizhevsky, I.~Sutskever, and G.~E. Hinton.
\newblock Imagenet classification with deep convolutional neural networks.
\newblock In {\em NeurIPS}, pages 1097--1105, 2012.

\bibitem[\protect\citeauthoryear{LeCun \bgroup \em et al.\egroup
  }{1989}]{lecun1989backpropagation}
Y.~LeCun, B.~Boser, J.~S. Denker, D.~Henderson, R.~E. Howard, W.~Hubbard, and
  L.~D. Jackel.
\newblock Backpropagation applied to handwritten zip code recognition.
\newblock {\em Neural computation}, 1(4):541--551, 1989.

\bibitem[\protect\citeauthoryear{Lin \bgroup \em et al.\egroup
  }{2017}]{DBLP:conf/iccv/LinGGHD17}
T.~Lin, P.~Goyal, R.~B. Girshick, K.~M. He, and P.~Doll{\'{a}}r.
\newblock Focal loss for dense object detection.
\newblock In {\em ICCV}, pages 2999--3007, 2017.

\bibitem[\protect\citeauthoryear{McCulloch and
  Pitts}{1990}]{mcculloch1990logical}
W.~S. McCulloch and W.~Pitts.
\newblock A logical calculus of the ideas immanent in nervous activity.
\newblock {\em Bulletin of mathematical biology}, 52(1-2):99--115, 1990.

\bibitem[\protect\citeauthoryear{Nair and Hinton}{2010}]{nair2010rectified}
V.~Nair and G.~E. Hinton.
\newblock Rectified linear units improve restricted boltzmann machines.
\newblock In {\em ICML}, pages 807--814, 2010.

\bibitem[\protect\citeauthoryear{Paszke \bgroup \em et al.\egroup
  }{2019}]{paszke2019pytorch}
A.~Paszke, S.~Gross, F.~Massa, A.~Lerer, J.~Bradbury, G.~Chanan, T.~Killeen,
  Z.~Lin, N.~Gimelshein, L.~Antiga, et~al.
\newblock Pytorch: An imperative style, high-performance deep learning library.
\newblock In {\em NeurIPS}, pages 8024--8035, 2019.

\bibitem[\protect\citeauthoryear{Ren \bgroup \em et al.\egroup
  }{2015}]{DBLP:conf/nips/RenHGS15}
S.~Q. Ren, K.~M. He, R.~B. Girshick, and J.~Sun.
\newblock Faster {R-CNN:} towards real-time object detection with region
  proposal networks.
\newblock In {\em NeurIPS}, pages 91--99, 2015.

\bibitem[\protect\citeauthoryear{Russakovsky \bgroup \em et al.\egroup
  }{2015}]{russakovsky2015imagenet}
O.~Russakovsky, J.~Deng, H.~Su, J.~Krause, S.~Satheesh, S.~Ma, Z.~H. Huang,
  A.~Karpathy, A.~Khosla, M.~Bernstein, et~al.
\newblock Imagenet large scale visual recognition challenge.
\newblock {\em International journal of computer vision}, 115(3):211--252,
  2015.

\bibitem[\protect\citeauthoryear{Selvaraju \bgroup \em et al.\egroup
  }{2017}]{DBLP:conf/iccv/SelvarajuCDVPB17}
R.~R. Selvaraju, M.~Cogswell, A.~Das, R.~Vedantam, D.~Parikh, and D.~Batra.
\newblock Grad-cam: Visual explanations from deep networks via gradient-based
  localization.
\newblock In {\em ICCV}, pages 618--626, 2017.

\bibitem[\protect\citeauthoryear{Sifre and Mallat}{2014}]{sifre2014rigid}
L.~Sifre and S.~Mallat.
\newblock Rigid-motion scattering for texture classification.
\newblock {\em arXiv:1403.1687}, 2014.

\bibitem[\protect\citeauthoryear{Simonyan and
  Zisserman}{2015}]{simonyan2014very}
K.~Simonyan and A.~Zisserman.
\newblock Very deep convolutional networks for large-scale image recognition.
\newblock In {\em ICLR}, 2015.

\bibitem[\protect\citeauthoryear{Tan and Le}{2019}]{DBLP:conf/icml/TanL19}
M.~X. Tan and Q.~V. Le.
\newblock Efficientnet: Rethinking model scaling for convolutional neural
  networks.
\newblock In {\em ICML}, volume~97, pages 6105--6114, 2019.

\bibitem[\protect\citeauthoryear{Wang \bgroup \em et al.\egroup
  }{2019}]{wang2019kervolutional}
C.~Wang, J.~F. Yang, L.~H. Xie, and J.~S. Yuan.
\newblock Kervolutional neural networks.
\newblock In {\em CVPR}, pages 31--40, 2019.

\bibitem[\protect\citeauthoryear{Xie \bgroup \em et al.\egroup
  }{2017}]{xie2017aggregated}
S.~N. Xie, R.~Girshick, P.~Doll{\'a}r, Z.~W. Tu, and K.~M. He.
\newblock Aggregated residual transformations for deep neural networks.
\newblock In {\em CVPR}, pages 1492--1500, 2017.

\bibitem[\protect\citeauthoryear{Xu \bgroup \em et al.\egroup
  }{2020}]{DBLP:conf/eccv/XuLLL20}
G.~D. Xu, Z.~W. Liu, X.~X. Li, and C.~C. Loy.
\newblock Knowledge distillation meets self-supervision.
\newblock In {\em ECCV}, volume 12354, pages 588--604, 2020.

\bibitem[\protect\citeauthoryear{Zoumpourlis \bgroup \em et al.\egroup
  }{2017}]{zoumpourlis2017non}
G.~Zoumpourlis, A.~Doumanoglou, N.~Vretos, and P.~Daras.
\newblock Non-linear convolution filters for cnn-based learning.
\newblock In {\em ICCV}, pages 4761--4769, 2017.

\end{thebibliography}

\end{document}